\documentclass{eptcs}
\providecommand{\event}{ADG 2021} 
\usepackage{graphicx}
\usepackage{hyperref}
\usepackage{xcolor}

\newcommand{\OGPCP}{\emph{Open Geometry Prover Community Project}}
\newcommand{\ogpcp}{\emph{OGPCP}}

\newcommand{\tgtp}{\emph{TGTP}}
\newcommand{\JGEX}{\emph{JGEX}}

\def\orcidID#1{\unskip$^{\mbox{\href{https://orcid.org/#1}{\scriptsize{[#1]}} }}$}

\title{Open Geometry Prover Community Project
\thanks{This work is funded
    by national funds through the FCT - Foundation for Science and
    Technology, I.P., within the scope of the project CISUC -
    UID/CEC/00326/2020 and by European Social Fund, through the
    Regional Operational Program Centro 2020.} 
} 

\author{Nuno Baeta\orcidID{0000-0002-1629-7924}
\institute{CISUC \\
University of Coimbra, Portugal} 
\email{\quad nmsbaeta@gmail.com} 
\and
Pedro Quaresma\orcidID{0000-0001-7728-4935}
\institute{CISUC, Department of Mathematics \\
University of Coimbra, Portugal}
\email{pedro@mat.uc.pt}
}

\date{}
\def\titlerunning{\OGPCP}
\def\authorrunning{Baeta \& Quaresma}

\begin{document}

\maketitle


\begin{abstract}
  Mathematical proof is undoubtedly the cornerstone of
  mathematics. The emergence, in the last years, of computing and
  reasoning tools, in particular automated geometry theorem provers,
  has enriched our experience with mathematics immensely. To avoid
  disparate efforts, the {\OGPCP} aims at the integration of the
  different efforts for the development of geometry automated theorem
  provers, under a common ``umbrella''. In this article the necessary
  steps to such integration are specified and the current
  implementation of some of those steps is described.
\end{abstract}

\setcounter{footnote}{0}

\section{Introduction}
\label{sec:introduction}

Mathematical proof is undoubtedly the cornerstone of mathematics. All
mathematics practitioner know its centrality and the difficulty in
mastering it~\cite{Hanna2019}.  The emergence, in the last years, of
computing and reasoning tools, in particular automated geometry theorem
provers, has enriched our experience with mathematics immensely.
Building such tools and exploring their applicability require a
coherent, well-organized community of researchers working in a
collaborative way, to avoid disparate efforts, as recalled by T.\ Han et
al.~\cite{Han2019}. Reuse of previous knowledge is vital for human
beings in all kinds of learning activities, and so much more in
mathematics.  The reuse of practical implementations of an abstract idea
is usually much harder than the reuse of the abstract idea itself.  The
same algorithm may be implemented several times using different
programming languages and data formats due to engineering mismatches.



The {\OGPCP} (\ogpcp) aims at the integration of the different efforts
for the development of geometry automated theorem provers, under a
common ``umbrella''.  As such, a contribution to the larger goal of
establishing a network of researchers working in the area of formal
reasoning, knowledge-based intelligent software and geometric knowledge
management, to explore efficient methodologies for the creation and reuse
of electronic tools in geometry.

To bring up such a framework a series of tools and protocols must
be implemented/established. The {\OGPCP} framework, goals are:

\begin{itemize}
\item to provide a common open access repository for the development
  of Geometry Automated Theorem Provers (GATP);
\item to provide an API to the different GATP in such a way that they
  can be easily used by users, stand-alone or integrated in other
  tools;
\item to develop portfolio strategies to allow choosing the best GATP
  for any given geometric conjecture;
\item to interface with repositories of geometric
  knowledge~\cite{Quaresma2018c} (e.g. \emph{TGTP}\footnote{Thousand
    of Geometric problems for geometric Theorem Provers,
    \url{http://hilbert.mat.uc.pt/TGTP/}}~\cite{Quaresma2011},
  \emph{TPTP}\footnote{ Thousands of Problems for Theorem Provers,
    \url{http://www.tptp.org/}}~\cite{Sutcliffe2017});
\item to develop a GATP System Competition to be able to rate
  GATPs~\cite{Baeta2020,Quaresma2019b}.
\end{itemize}

\paragraph{Overview of the paper.} The paper is organised as follows:
first, in \S\ref{sec:ogpstatus}, the current status of the framework
implementation is described. In \S\ref{sec:gatps} a short description
of the GATP currently incorporated in the {\ogpcp} is given. Finally,
in \S\ref{sec:conclusions}, conclusions are drawn and future work is
discussed.

\section{{\ogpcp} Implementation Status}
\label{sec:ogpstatus}

The {\ogpcp} framework is a never-ending project in the sense that new
GATP can be proposed and incorporated in the project at any given
moment. Nevertheless many of the steps necessary for its current use
and for an easy integration of new future projects are already done.

\begin{figure}[htb!]
    \centering
    \includegraphics[width=0.9\textwidth]{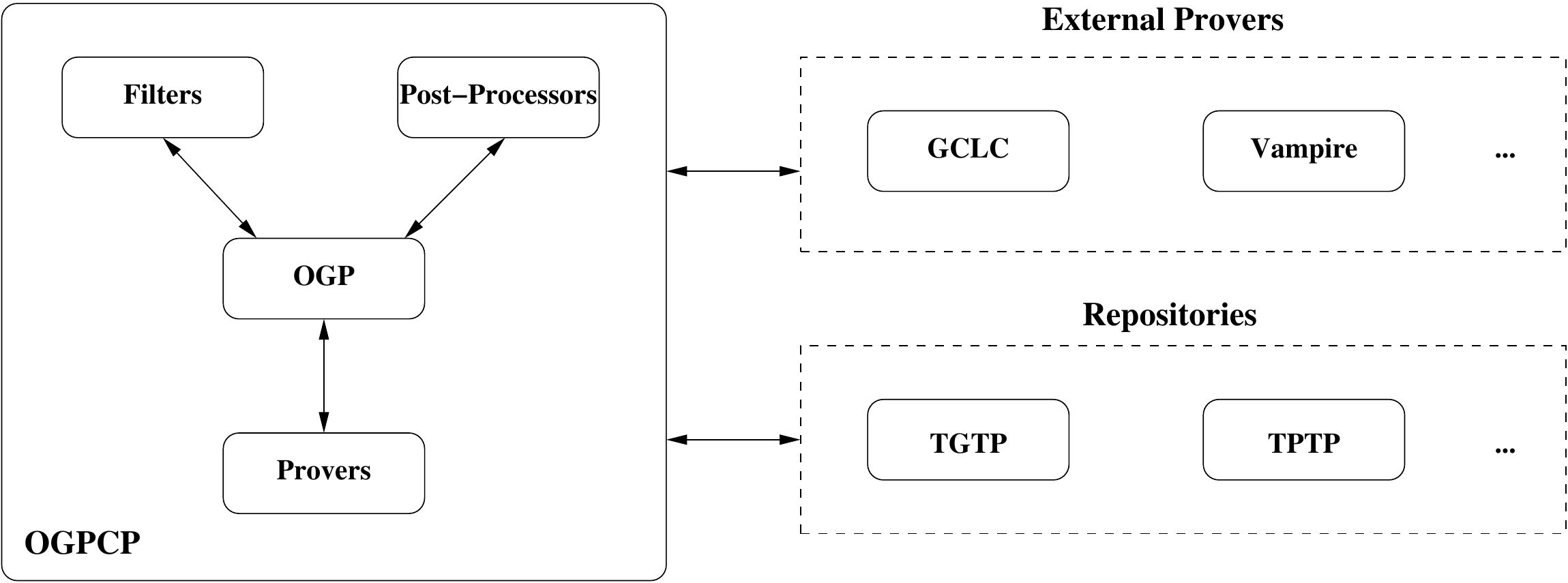}
    \caption{OGPCP Framework}
    \label{fig:ogpcpframework}
\end{figure}


\subsection{{\ogpcp} Source Repository}
\label{sec:ogpcprepository}

The {\ogpcp} is hosted at
GitHub.\footnote{\url{https://github.com/opengeometryprover/OpenGeometryProver}}
The code is made available under the GNU General Public
License,\footnote{\url{https://www.gnu.org/licenses/gpl.html}} version
3 or later, and the documentation under the GNU Free Documentation
License,\footnote{\url{https://www.gnu.org/licenses/fdl.html}} version
1 or later.

{\ogpcp} is available only as source-code and its installation in Unix systems
is a straightforward process, provided that GNU Make, Apache Ant and OpenJDK
are installed.  After downloading the code from the GitHub repository, in the
command line just type
\begin{quote}
    \texttt{\$ make} \\
    \texttt{\$ make install} \\
\end{quote}


\subsection{{\ogpcp} Application Programming Interface}
\label{sec:ogpcpapi}

{\ogpcp} API is a combination of several command-line tools, e.g.,
native (i.e. done by the {\ogpcp} team and sharing a common base code) 
and external provers, filters, post-processors, prepared to seamlessly work
together or with independent tools.

Native {\ogpcp} provers must:
\begin{itemize}
\item use \emph{TPTP}'s first-order format (FOF) syntax as their
  default conjecture format; 
\item accept the same command-line arguments;
\item provide the same output;
\end{itemize}

All this is explained in \textsl{\OGPCP\ Programmer Manual}, available
at the {\ogpcp}'s GitHub repository.

External provers will be developed by other teams, with different base
codes, even, eventually, using different programming languages (no
enforcement is done on those matters). External provers conjecture's
format may not adhere to TPTP's FOF syntax (see
Section~\ref{sec:conclusions}).  In such cases, for {\ogpcp} to take
advantage of those provers, and vice-versa, filters to/from the FOF
syntax must be written, as well as post-processors to interpret the
output of those provers.

\medskip
Example of usage and already implemented features:

\begin{description}
\item[Using conjectures \emph{in situ}, contained in local files] {\ }
  
  \begin{description}
  \item[\texttt{\$ ogp ceva.gcl}] \hfill use of GCL prover, native
    language, area method 
  \item[\texttt{\$ ogp ceva.gcl -w}] \hfill use of GCL prover, native
    language, Wu's method
  \item[\texttt{\$ ogp ceva.coqam}] \hfill use of CoqAM, native
    language, area method
  \item[\texttt{\$ ogp ceva.fof}] \hfill use of inbuilt portfolio mechanism
  \item[\texttt{\$ ogp -t 30 ceva.fof }] \hfill use of inbuilt portfolio
    mechanism with a time limit (30 seconds)
  \end{description}
\item[Using conjectures in a remote repository] {\ }
  \begin{description}
  \item[\texttt{\$ ogp --tgtp=GEO0001 gclc}] \hfill connection to the
    \emph{TGTP} repository (see §~\ref{sec:ogcpinterfacerepositories}).
  \end{description}
\end{description}

The command line {\ogpcp} meta-syntax is the following:



\begin{quote}
  \texttt{ogp [$<$option$>$] [<conjecture> [<prover> [<prover-options>]]}]
\end{quote}

The available options are:
\begin{description}
\item[\texttt{-h}, \texttt{--help}] \ \\
  prints a help message and exits --- to be used alone;
\item[\texttt{-p}, \texttt{--provers}] \ \\
  lists the available (to {\ogpcp}) GATP and exits --- to be used alone;
\item[\texttt{-V}, \texttt{--version}] \ \\
  prints {\ogpcp} version and exits --- to be used alone;
\item[\texttt{-t $<$time$>$}, \texttt{--timeout=$<$time$>$}] \ \\
  redefines the default time limit, in seconds, when proving a
  conjecture.
\end{description}
The conjecture is provided to the prover in a local file or, when
using the remote repository \emph{TGTP}, by its unique id
in the repository, using the syntax
\begin{center}
    \texttt{--tgtp=$<$conjecture\_id$>$}.
\end{center}
When attempting to prove a conjecture, the choice of the prover, if
none is indicated, proceeds according to the following rules:
\begin{enumerate}
\item if the conjecture is provided in a local file, then
  \begin{enumerate}
  \item if file name extension is \texttt{fof}, then use {\ogpcp}
    portfolio prover;
  \item otherwise, use the default prover associated with that
    extension
  \end{enumerate}
\item otherwise, use {\ogpcp} portfolio prover.
\end{enumerate}
When the prover is given, a check is made to certify if the conjecture's
format is one accepted by the prover.  If that is not the case, whenever
possible filters are used to convert to a format accepted by the prover.
If all this fails, an error occurs and the process ended.


\subsection{{\ogpcp} Filters \& Post-processors}
\label{sec:filterspostprocessors}

A set of filters are already ready to be used.

\begin{description}
\item[\texttt{filterGCLtoFOF}] \hfill GCL language to FOF
\item[\texttt{filterGEOGEBRAtoFOF}] \hfill
  GeoGebra\footnote{\url{https://www.geogebra.org/}} to FOF
\item[\texttt{filterJGEXtoFOF}] \hfill JGEX to FOF
\end{description}
for the moment all these filters, \texttt{filter$\ast$toFOF}, assume
the inclusion of the axioms of the deductive database full-angle
method~\cite{Chou2000}, given that these are already converted to FOF
syntax.\footnote{GEO012+0.ax, 
  \url{http://www.tptp.org/cgi-bin/SeeTPTP?Category=Axioms}} That is,
a plain conversion is made and an \texttt{include} instruction is
added at the begin with the above mentioned axiom set.




Post-processors are to be used in conjunction with independent
provers.  They are used to obtain information about the proof's
result, e.g., if the proof was successful or not, time, file with the
proof steps, if any, etc., as the output of an independent prover must
not adhere to that of a native {\ogpcp} prover.

As of this writing, there is only one post-processor --- for the
Vampire ATP, to get the time of a proof.


\subsection{{\ogpcp} Portfolio}
\label{sec:ogpcpportofolio}

Portfolio problem solving is an approach in which for an individual
instance of a specific problem, one particular, hopefully the most
appropriate, solving technique is automatically selected among several
available ones and used. Weindenbach~\cite{Weidenbach2017} makes the
distinction between syntactic and semantic approaches. With a
\emph{Simple-Syntactic portfolio solver} the selection of the core
solvers is done by purely syntactic problem properties and there is no
exchange of results between different core solvers. In a
\emph{Sophisticated-Semantic portfolio solver} the selection of the
core solvers is done by semantic or structural problem properties and
the solvers exchange results~\cite{Weidenbach2017}.

Already some work in the area of geometric automated theorem proving
has been done, namely in the prover mechanism implemented in
GeoGebra~\cite{Kovacs2014,Kovacs2015,Nikolic2018}. It is expected that
this research can be incorporated into the {\ogpcp}.

\subsection{{\ogpcp} Interface with Repositories}
\label{sec:ogcpinterfacerepositories}

A server/client architecture to connect {\ogpcp} and {\tgtp} is
already available. On the side of the {\tgtp} repository a query-server is
already implemented, always listening to client requests.

The code for the clients is open-source and available as part of the
{\ogpcp} project. The clients are build in such a way that a SQL query
can be send to the {\tgtp} database, receiving in return the code of
the desired conjecture. The exchange of information between the server
and the client is done using the
\emph{JSON} format.\footnote{\url{https://www.json.org/}} The
implementation of new clients to other GATP it is easy and opens the
use of the information contained in {\tgtp} from any GATP. This
server/client architecture is currently being used to establish a
connection between the e-learning environment \emph{Web Geometry
  Laboratory} and the {\tgtp}
repository~\cite{Quaresma2018c,Quaresma2018a}.

For example, using the {\ogpcp} API we could write: \texttt{ogp
  --tgtp=GEO0001 gclc}. This call will trigger the \texttt{tgtpToOgpcp}
client, sending a query about the \emph{gclc} code for problem
\emph{GEO0001} in {\tgtp}. If the problem do not exist an error code
will be returned, if the \emph{gclc} for such a problem do not exist,
the \emph{FOF} code for that problem will be returned. After receiving
an error free answer, the \texttt{ogp} command will pursue as usual.


\subsection{Geometry Automated Theorem Provers Systems
  Competition}
\label{sec:gasc}

To be able to compare the different methods and implementations, a
competition will have the virtue of pushing towards the standardization
of the input language, the standardization of test sets, the direct
comparability and the easier exchange of ideas and algorithmic
techniques. The results of such a competition will also constitute a
showcase, where potential users will look for the best GATP for
their goals~\cite{Baeta2020,Quaresma2019b}.

A first trial-run of the \emph{Geometry Automated Theorem Provers
  Systems Competition}, GASC 0.2, was already run, at ThEdu’19, the
\emph{8th International Workshop on Theorem proving components for
  Educational software}, August 2019, Natal,
Brazil~\cite{Quaresma2019b,Quaresma2020c} and a second trial-run is
being prepared.

Not being directly related to the {\ogpcp} the GASC will be used to
test the different GATP in the project, pushing towards the
development of new and better implementations.



\section{External Geometry Automated Theorem Provers}
\label{sec:gatps}

A set of external GATP are already part of the OGPCP. These are
autonomous open source projects that recognise the {\ogpcp} and from
which filters to/from the native syntax and FOF are already
implemented, or will be implemented in a near future.

Those GATP must be downloaded and installed in a separate way, simple
instructions on how to do it will be part of the {\ogpcp}
documentation.




\medskip

\paragraph{GeoGebra Automated Reasoning Tools.}
\label{sec:GeoGebra} The standard version of
\emph{GeoGebra}\footnote{\url{https://www.geogebra.org/download}}
includes several Automated Reasoning Tools (ART):

\begin{itemize}
\item for conjecturing a geometric property (e.g.~such three points
  visually ``seem'' to be aligned), the \texttt{Relation} command;
\item for rigorously denying or confirming a given conjecture
  (e.g.~providing an affirmative answer to the conjecture after
  internally verifying, using Computer Algebra tools, that some
  determinant involving the coordinates of the three selected points
  is zero), the \texttt{Prove} and \texttt{ProveDetails} commands;
\item for presenting some complementary hypotheses for the truth of a
  given (actually false) statement (e.g.~remarking that the truth of
  the proposed statement needs some further steps in the geometric
  construction describing the statement), the \texttt{LocusEquation}
  command.
\end{itemize} 

See~\cite{Botana2015a} for a detailed explanation about the project
and~\cite{Kovacs2018} for a tutorial-like paper about the different
commands, as well as~\cite{Zoltan2022} for a quite updated version.

Moreover there are two other reasoning toolsets, already implemented
but in (yet) non-standard versions of
\emph{GeoGebra}~\cite{Botana2020,Kovacs2020a}. The first one contains
the \texttt{Discover} tool and command, and the \texttt{Compare}
command, can be used in the \emph{GeoGebra Discovery}
fork,\footnote{\url{https://github.com/kovzol/geogebra-discovery},
  \url{http://autgeo.online/geogebra-discovery/}} available in two
different options: one, operating over \emph{GeoGebra Classic 5}, for
\emph{MS-Windows}, \emph{Mac} and \emph{Linux} systems; and, the
other, working over \emph{GeoGebra Classic 6}, that requires starting
it in a browser, for tablets and smartphones.

The \texttt{Discover} command automatically finds all theorems (of a
certain kind: parallelism, congruence, perpendicularity, etc.) holding
over a given element of a construction (e.g.~involving a point), by
considering some combinatorial heuristics to formulate different
\emph{Relation} tests involving always the selected element, plus some
other one, and presenting as output the collection of obtained
properties. The \texttt{Compare} command is used to find a general
relationship between two quantities (for example, by comparing the sum
of the lengths of the catheti $a$ and $b$ and the length of the
hypotenuse $c$ in a right triangle---clearly, here the relationship
is an inequality, namely, $c<a+b<\sqrt2 c$). This low-level command is
usually called from an improved version of the \texttt{Relation}
command \cite{VajdaKovacs2020}.

The second currently on-going improvement deals with the development
of an \emph{AG=Automated Geometer},\footnote{\url{http://autgeo.online/ag/}, \url{https://github.com/kovzol/ag}}
a ``geometer'' that does not require human intervention, except that
of launching the computation process over a figure.  It is a web-based
module that allows GeoGebra to automatically produce different
conjectures over the given geometric construction, and to internally
confirm or deny them using tools similar to those in GeoGebra
Discovery, but here not limited to exploring relations involving a
single, specific element.

The algorithms behind all these tools deal with the algebraic
translation of the geometric statements and the symbolic
manipulation---via the embedded computer algebra system
GIAC\footnote{Giac/Xcas, \url{https://www-fourier.ujf-grenoble.fr/~parisse/giac.html}}~\cite{Kovacs2015b}---of
the corresponding complex algebraic geometry
varieties. See~\cite{Kovacs2019,Kovacs2019a,Ladra2020,Recio1999}, for
a detailed description of the involved theoretical approach.

It must be remarked that the chosen method is quite effective and is
able to deal, in milliseconds and over a variety of popular electronic
devices (laptops, smartphones, etc.), with very complicated statements
but, on the other hand, it does not provide any human-understandable
arguments for the declared truth/falsity of the involved statements.

Finally, we summarize how GeoGebra's features can be directly
exploited by OGPCP. GeoGebra offers two application programming
interfaces for external programs:
\begin{itemize}
\item A JavaScript Application Programming Interface (API), available
  for web applications. The suggested method is to set up the
  construction via JavaScript calls and then execute the command
  \texttt{ProveDetails} to obtain the result.
\item The desktop application can be directly called with an input
  GeoGebra file. The file structure is given in XML. By creating the
  XML data as input, and calling GeoGebra via command line, the debug
  information can be directly processed to get the result.
\end{itemize}

\paragraph{GCLC Automated Reasoning Tools.}
\label{sec:GCLC}
Within the mathematical software GCLC, there is an implementation of
the area method~\cite{Janicic2012a} by Jani\v{c}i\'c and Quaresma, and
implementations of (simple) Wu's method and Gr\"obner based method, by
Predovi\'c and Jani\v{c}i\'c~\cite{Janicic2006c}. Apart a graphical
user interface all the GATP can be used in stand-alone mode, begin
usable in the overall \OGPCP\ interface.

\paragraph{CoqAM.} 
\label{sec:CoqAM} The formalization within the \emph{Coq} proof
assistant of the area method, a decision procedure for affine plane
geometry~\cite{Janicic2012a,Narboux2004}.\footnote{\url{https://dpt-info.u-strasbg.fr/~narboux/area_method.html}}
It can be used in stand-alone mode (\emph{Coq} must be installed),
being possible its use within overall \OGPCP\ interface.


\paragraph{JGEX.}
\label{sec:jgex} Java Geometry Expert, {\JGEX}, is a software which
combines dynamic geometry software (DGS), automated geometry theorem
prover (GATP) and an approach for visually dynamic presentation of
proofs. As a dynamic geometry software, {\JGEX} can be used to build
dynamic visual models to assist teaching and learning of various
mathematical concepts.\footnote{\url{https://github.com/yezheng1981/Java-Geometry-Expert}}

Apart the use of the GATP systems inside the overall graphical DGS
interface, they can be used stand-alone, being possible its use
within the overall \OGPCP\ interface.


\paragraph{Generic ATP}
\label{sec:genericATP}

Apart from these ATP, specific to geometry (GATP), the generic ATP can
also be used. It is a question of including an axiomatic theory
specific to geometry, e.g. those in the \emph{Geo} domain in
\emph{TPTP} (Hilbert geometry; Tarski geometry, Rules of construction
(von Plato), deductive database method, among others). Not being in
the core of the project its use is, nevertheless, being taken in
consideration and the \OGPCP\ command line tool will process an input
related to such tools.


\section{Conclusions and Future Work}
\label{sec:conclusions}

The problems related to the integration between different geometry
provers can be much more harder than the presented above. Different
algorithms/provers do not assume all the same mathematical setting.
Different axiomatizations exist, e.g. Tarski's, Hilbert’s, von
Plato's; Area method. Different kinds of geometry, e.g. euclidean 2D
or 3D, non-euclidean. Different types of approaches, geometric,
e.g. area method, algebraic, e.g. Wu's method. More than a, maybe
unrealistic, full integration, the {\ogpcp} should aim to: give a
simple, documented, open source, API to allow the use of GATP by
experts and non-experts and to cons\-titute itself as a forum, a space
of discussion, about the deductive tools for geometry.


Apart many improvements in the existing framework, e.g. improve the API,
linking with external provers, filters and post-processing, new native
provers are planned: a new implementation of the full-angle
method~\cite{chou1996d}, the deductive database method~\cite{Chou2000}
(using the axioms of the full-angle method), and a novel approach, the
deductive graphs method, based on the deductive database method but
using deductive graphs. Some initial work has already been done in those
methods~\cite{Baeta2013,Haralambous2014,Haralambous2018}.

As said at the beginning the {\ogpcp} is meant to be a never ending
project in the sense that new improvements in the area of automated
deduction will be made and incorporated in it. New methods, new
implementations, improvements in the existing approaches, etc. To
enlarge the usefulness and conquer new ``audiences'' (e.g. teachers in
primary and secondary tools) the GATP need to be more modular, being
able to be incorporated into ``friendly'' tools, that can cover the
``difficult nature'' of many GATP. The {\ogpcp} should help on the
goal of ``bring the automated deduction to all geometers''.


\subsubsection*{Acknowledgements} Zolt{\'a}n Kov{\'a}cs, Tom{\'a}s
Recio, Francisco Botana (\emph{GeoGebra}), Predrag Jani{\v c}i{\'c}
(\emph{GCLC}) and Julien Narboux (\emph{COQAm}), contributed to this
work by helping writing section~\ref{sec:gatps}.

\bibliographystyle{eptcs}


\bibliography{2021_NBPQ_ADG21_OGPCP}

\end{document}